\documentclass[11pt,a4paper,onecolumn,oneside]{article}
\usepackage[T1]{fontenc}
\usepackage[utf8]{inputenc}
\usepackage{lmodern}
\usepackage[a4paper,margin=25mm]{geometry}
\usepackage{microtype}
\usepackage{amsmath,amssymb}
\usepackage{array,booktabs,tabularx,longtable}
\usepackage{float,flafter,placeins,graphicx}
\usepackage{tikz}
\usetikzlibrary{arrows.meta,positioning,calc}
\usepackage{setspace}
\usepackage[font=small,labelfont=bf,skip=5pt]{caption}
\usepackage[numbers,sort&compress]{natbib}
\usepackage{xurl}
\usepackage[hidelinks]{hyperref}
\newcommand{\PreprintAuthor}[2]{#1\textsuperscript{#2}}
\newcommand{\PreprintAffiliation}[2]{\textsuperscript{#1}#2\par}
\newcommand{\PreprintFrontMatter}[4]{%
  \begin{center}
    {\LARGE\bfseries #1\par}\vspace{.8em}
    {\large #2\par}\vspace{.55em}
    {\small #3}\vspace{.4em}
    {\small\textsuperscript{*}Corresponding author: #4\par}
  \end{center}
}
\newcommand{\PreprintKeywords}[1]{\noindent\textbf{Keywords:} #1\par}
\newcommand{\PreprintBibliography}[1]{%
  \begingroup\raggedright
  \bibliographystyle{unsrtnat}%
  \bibliography{#1}%
  \endgroup
}

\hypersetup{
  pdftitle={Benchmarking EEG feature families for ADHD research in children and adolescents: performance and explainability across attention tasks},
  pdfauthor={Javier Sanchis, Jorge García-Carrasco, Álvaro Navarro, Tahar Kechadi, Juan Trujillo}
}

\begin{document}
\PreprintFrontMatter
  {Benchmarking EEG feature families for ADHD research in children and adolescents: performance and explainability across attention tasks}
  {\PreprintAuthor{Javier~Sanchis}{1,*},
   \PreprintAuthor{Jorge~Garc\'ia-Carrasco}{1},
   \PreprintAuthor{\'Alvaro~Navarro}{1},
   \PreprintAuthor{Tahar~Kechadi}{2},
   \PreprintAuthor{Juan~Trujillo}{1}}
  {\PreprintAffiliation{1}{Lucentia Research, Department of Software and Computing Systems, University of Alicante, Sant Vicent del Raspeig, Spain}
   \PreprintAffiliation{2}{School of Computer Science and Informatics, University College Dublin, Dublin, Ireland}}
  {Javier Sanchis, \href{mailto:javier.sanchis@ua.es}{javier.sanchis@ua.es}}

\begin{abstract}
Attention Deficit Hyperactivity Disorder varies in presentation.
Diagnosis requires clinical and developmental assessment.
Reproducible electroencephalography measures may complement clinical and behavioural characterisation, but differing feature definitions and evaluation methods hinder comparison.
We compared seven feature families and four combinations across attention tasks and assessed the stability of held-out explanations.
Two serious games yielded four overlapping attention-task cohorts of 123-140 participants each.
Repeated nested participant-held-out evaluation assessed recruitment-group discrimination.
Exploratory paired participant-cluster bootstrap tests corrected for multiplicity in two separate families: 72 individual-task contrasts and 18 fusion contrasts.
For eight configurations selected after inspecting predictive results, we explained held-out native probabilities before calibration.
In the strongest task context, the highest configuration mean AUROC over three repetitions was 0.773, with a fixed-prediction 95\% interval of 0.692-0.854 unadjusted for selecting the maximum.
Precision, sensitivity and specificity were 0.740, 0.556 and 0.861.
In that task, the paired AUROC difference for connectivity minus spectral features averaged +0.2083 across six classifiers and the common 8 s and 16 s durations, with pointwise 95\% CI 0.1076-0.3072 and adjusted $p = 0.0036$.
F4-AF4 beta-band coherence led repetition-level attribution summaries in five selected procedures and was selected in every fit.
Other contexts showed weaker discrimination and less consistent retention of leading scientific inputs.
No general classifier winner emerged.
Connectivity and its recurring coherence input warrant independent replication, but do not justify a universal feature recommendation.
These exploratory findings do not establish causal task effects, disease mechanisms or diagnostic validity.
\end{abstract}

\PreprintKeywords{EEG, ADHD, Feature selection, Benchmarking, Explainability, Connectivity}

\section{Introduction}
Attention Deficit Hyperactivity Disorder (ADHD) is a heterogeneous neurodevelopmental condition \cite{biomarkers}. Diagnosis relies on clinical and developmental assessment, including symptoms and functioning across everyday settings \cite{nice2018adhd}. This heterogeneity motivates research into reproducible neurophysiological measurements that might complement clinical and behavioural characterisation \cite{lohani2025machine}. Electroencephalography (EEG) records neural activity non-invasively with high temporal resolution \cite{lyu2026eeg}, but the information extracted depends on how the signal is represented. Comparing representations under common analytical conditions can therefore help identify measurements worth further investigation. Clinical application nevertheless requires independent replication and validation of individual-level performance. These candidate measurements do not replace clinical assessment \cite{biomarkers}.

Feature extraction turns these recordings into quantities that can be compared across participants, while preserving different aspects of the signal. Temporal/statistical features summarise amplitude and variability, whereas conventional spectral features describe band power and ratios \cite{features}. Aperiodic spectral features distinguish broadband background structure from oscillatory peaks \cite{donoghue2020}, and complexity/fractal measures characterise irregularity and scaling \cite{features}. To describe interactions, functional connectivity quantifies statistical coupling and graph/network features summarise the organisation of those connections \cite{ghaderi2025altered}. Time-frequency decompositions describe structure across temporal and frequency scales \cite{lyu2026eeg}. These seven families offer different measurement perspectives, but their definitions alone do not establish independent predictive information.

Recent work combines multi-domain measurements with explainable, stability-aware selection \cite{mohammedatheef2026mrsshap}. However, favourable scores across studies remain difficult to interpret because cohorts, acquisition, preprocessing, feature definitions, durations, models and evaluation designs also differ \cite{quinterolopez2025review}. A score reflects that whole analytical setting, rather than the feature family alone. What remains unclear is which feature families and combinations show favourable discrimination under a common participant-level evaluation protocol, and how consistently these patterns recur across task contexts.

Those comparisons are informative only if evaluation limits optimistic bias. When correlated epochs from one participant occur in both training and test sets, apparent performance on unseen participants can instead reflect participant-specific information \cite{brookshire2024eegleakage}. Keeping participants separate addresses this leakage, but feature selection and tuning must also be isolated from final evaluation through nested validation \cite{varma}. The safeguards serve different purposes: grouping separates people, while nesting separates model choice from performance estimation. Neither turns additional epochs into additional independent participants. Small samples can consequently leave substantial uncertainty \cite{varoquaux2018crossvalidation}, and internal validation does not establish performance in an external population.

To address this gap, we draw on public BALLADEER EEG recordings collected during two serious games, Attention Robots and Attention Slackline \cite{balladeer,source}. Participants selected robots that matched target features in Robots and pressed a button in response to on-screen flag patterns in Slackline \cite{robots,slackline}. We treat four recording contexts as separate tasks in this benchmark, one for Robots and one for each of the three recorded Slackline levels (L1, L6 and L11). Building on the existing EEG preparation and feature tables, we compare representations, supported durations and classifiers through participant-grouped nested evaluation within each task. Exploratory paired comparisons complement descriptive performance patterns. Held-out explanations of models selected after predictive inspection then examine which individual measurements contribute to their predictions.

The outcome is recruitment group, with participant-level AUROC as the primary criterion. Because cohorts, devices and partitions differ, comparisons between tasks remain descriptive. Four questions organise the existing exploratory study, rather than a preregistered protocol:
\begin{enumerate}
\renewcommand{\labelenumi}{RQ\arabic{enumi}.}
\interlinepenalty=10000
\item How do feature families and combinations compare in participant-level discrimination within each task?
\item How do observed performance and leading representation patterns vary across task contexts?
\item How do supported epoch durations and classifier choices relate to discrimination within each task?
\item Which inputs recur in the selected held-out explanations, and how consistent are their selection and attribution patterns?
\end{enumerate}

The study makes three contributions. First, it provides comparative evidence for seven families and four combinations across four attention-task contexts, supported durations and six classifiers under a common within-task participant-held-out protocol. Second, it identifies representation and duration choices for replication while showing why stronger Robots findings and weaker Slackline evidence do not support a universal task or classifier ranking. Third, it links broad representation findings to candidate individual measurements through held-out attribution and repeated selection in selected fitted models. Reporting both contribution and retention consistency helps distinguish recurring inputs from those prominent only in particular fits.

\section{Related Work}
EEG feature studies in ADHD have extended conventional spectral measurements with broader engineered representations \cite{features,quinterolopez2025review}. In the study reported in \cite{safi2026amplitude}, the authors compared conventional descriptors with a combined set that added amplitude-modulation descriptors, both extracted from wavelet-decomposed frequency bands. They reported improved classification across the tested classifiers and wavelet choices when modulation information was added. This result supports examining complementary descriptors within that pipeline, rather than ranking feature families across studies. The study in \cite{ghaderi2025altered} examined a different representation: source-localised connectivity characterised through network modularity and communication measures. These source-space networks differ from scalp-channel coherence. The studies therefore address complementary questions about what information a representation captures and whether it improves prediction.

Broader representations also raise the question of which inputs to retain. The MRS-SHAP framework presented in \cite{mohammedatheef2026mrsshap} combines attribution-based importance, temporal stability and performance-aware weighting to select feature subsets. Selection of individual inputs and comparison of feature families can coexist, but they answer different questions. This benchmark uses supervised selection within training folds for each specified feature view. Its held-out explanations then examine selected configurations after predictive inspection. Attribution describes contributions to those fitted predictions, alongside input retention across fits, rather than choosing the benchmark's feature subsets.

Interpreting such comparisons requires attention to both evaluation and temporal context. The analysis in \cite{brookshire2024eegleakage} demonstrated segment-partitioning leakage in Alzheimer's disease and epilepsy EEG. A separate study \cite{delpup2025partitioning} examined partitioning strategies for cross-participant prediction. These findings support keeping participants separate when comparing representations. In \cite{garciaponsoda2024refining}, the authors compared preprocessing and chronological recording segments, reporting stronger classification in later portions. Recording position concerns when information is acquired, whereas epoch duration determines the window used to calculate features. A benefit in later recording portions therefore does not establish which feature window is preferable.

Together, these precedents motivate comparisons that specify the representation, input selection, temporal unit and evaluation conditions. The present benchmark compares families and combinations across supported durations and classifiers under a common within-task protocol. It adds descriptive evidence on recurring patterns across task contexts and uses selected held-out explanations to identify measurements for replication. Differing cohorts, devices and partitions prevent those patterns from isolating causal task effects or establishing a universal feature winner.

\section{Methods}
\begin{figure}[htbp]
\centering
\begingroup
\linespread{1}\selectfont
\definecolor{wfblue}{HTML}{24556B}
\definecolor{wfpale}{HTML}{F0F5F7}
\definecolor{wfgreen}{HTML}{276154}
\definecolor{wflightgreen}{HTML}{EDF5F1}
\begin{tikzpicture}[
  node distance=4mm,
  font=\sffamily\small,
  stage/.style={draw=wfblue,fill=wfpale,line width=0.7pt,
    rounded corners=1.5mm,align=center,text width=.91\linewidth,
    inner xsep=3mm,inner ysep=2.5mm},
  output/.style={stage,text width=.405\linewidth,minimum height=29mm},
  flow/.style={-{Stealth[length=2mm]},draw=wfblue,line width=0.9pt}
]
\node[stage] (source) {\textbf{1. Source EEG and task cohorts}\\
  BALLADEER: Robots and Slackline L1/L6/L11\\
  Separate complete-case Slackline fusion cohort};
\node[stage,below=of source] (signal) {\textbf{2. Outcome-blind signal preparation}\\
  Filtering, resampling and gap-aware epochs};
\node[stage,below=of signal] (features) {\textbf{3. Fixed feature extraction and views}\\
  Seven feature families and four combinations};
\node[stage,below=of features,minimum height=43mm] (evaluation) {};
\node[align=center,anchor=north,font=\sffamily\small]
  at ([yshift=-2.5mm]evaluation.north)
  {\textbf{4. Nested participant-held-out evaluation}\\
   3 repetitions, 5 outer folds, 4 inner folds, 6 classifiers};
\node[draw=wfgreen,fill=wflightgreen,rounded corners=1mm,
  line width=0.7pt,text width=.81\linewidth,align=center,inner sep=2.5mm]
  at ([yshift=-1mm]evaluation.center)
  {\textbf{Training participants only}\\
   Learned preparation, selection, tuning and sigmoid fitting\\
   Refit preparation and classifier on outer-training participants};
\node[anchor=south,align=center,font=\sffamily\small]
  at ([yshift=2mm]evaluation.south)
  {Outer-test participants remain held out};
\node[output,anchor=north west] (predict)
  at ([yshift=-7mm]evaluation.south west)
  {\textbf{5a. Predictive evidence}\\
   Calibrated participant predictions\\
   Descriptive and within-cohort paired comparisons};
\node[output,anchor=north east] (explain)
  at ([yshift=-7mm]evaluation.south east)
  {\textbf{5b. Selected explanations}\\
   Post-results selection of tree fits\\
   Held-out attribution on native probability, before calibration};
\draw[flow] (source) -- (signal);
\draw[flow] (signal) -- (features);
\draw[flow] (features) -- (evaluation);
\draw[flow] (evaluation.south -| predict.north) -- (predict.north);
\draw[flow] (evaluation.south -| explain.north) -- (explain.north);
\end{tikzpicture}
\endgroup
\caption{From source EEG to predictive evidence and selected-model explanations.}
\label{fig:workflow}
\end{figure}
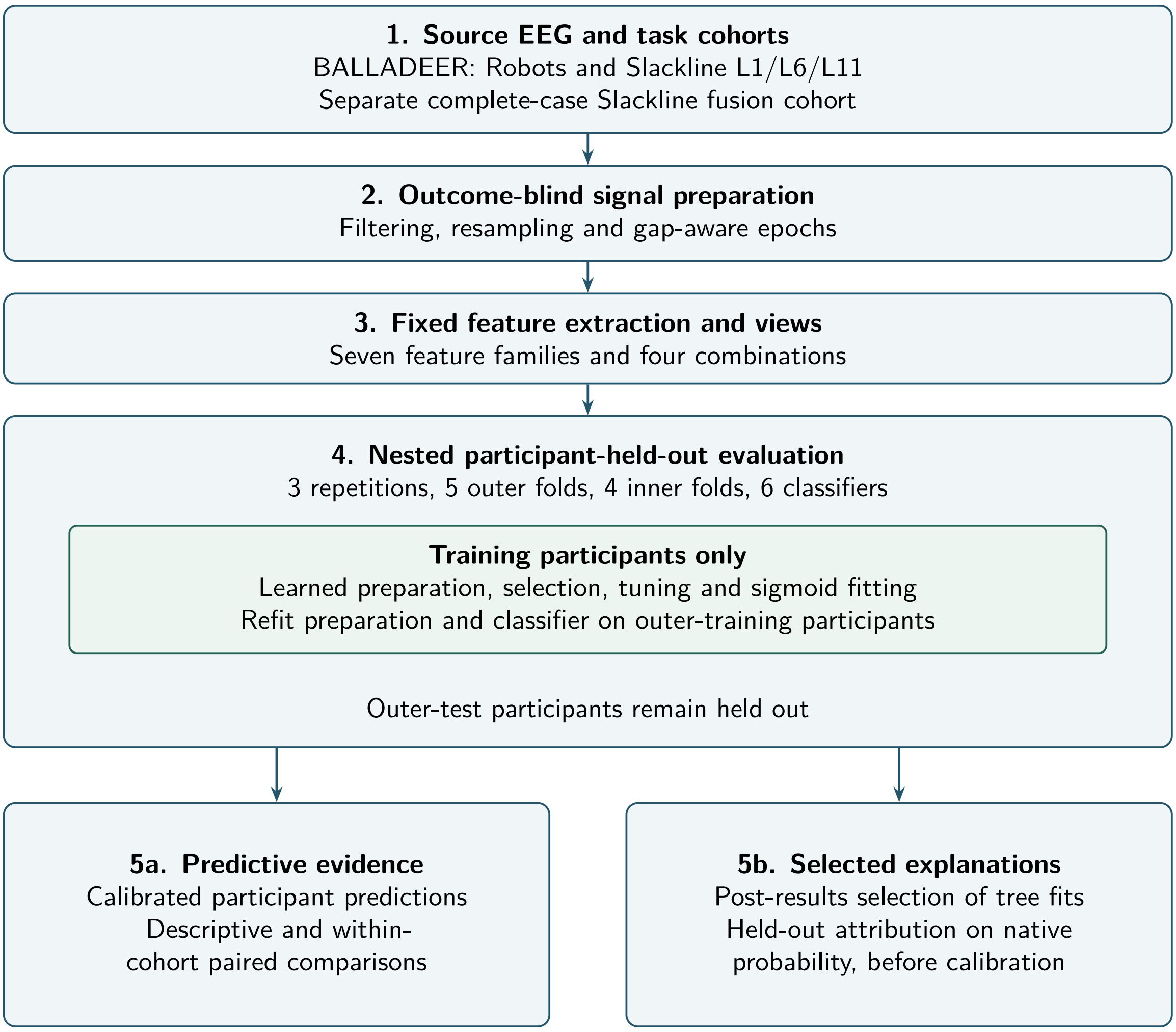

The analysis compared feature views, durations and classifiers for within-task recruitment-group discrimination. Participant-held-out predictions supported performance comparisons, while selected-model explanations examined recurring inputs. The workflow distinguishes inherited outcome-blind preparation from training-fold operations (Figure~\ref{fig:workflow}).

\FloatBarrier

\subsection{Data, task cohorts and inherited preprocessing}
We used our curated EEG and feature dataset \cite{sanchis2026taskeegdata}, derived from public BALLADEER recordings \cite{balladeer,source}. It contains preprocessed EEG, extracted features and documentation of data quality and processing. Across the full dataset, outcome-blind assessment retained 669 processed task recordings from 156 of 158 public participants aged 6-18 years across five source conditions. The benchmark evaluated Robots and each of the three Slackline levels as separate task contexts.

Processed task recordings and eligible participants are distinguished in Table~\ref{tab:cohorts}. Task cohorts overlap. For the separate fusion analysis, we used the complete-case subset of participants with recordings at all three Slackline levels, without imputing missing tasks. ADHD recruitment defined the positive class, separately from assessment-derived status. Robots used 14-channel Emotiv recordings and Slackline 29 retained CGX scalp channels.

\begin{table}[htbp]
\centering
\small
\caption{Recording conditions and predictive cohorts.}
\label{tab:cohorts}
\begin{tabularx}{\linewidth}{@{}>{\raggedright\arraybackslash}X rrr@{}}
\toprule
Condition (source code) & \shortstack{Processed\\recordings} & Participants & ADHD/comparison \\
\midrule
Robots (BAL02) & 123 & 123 & 51/72 \\
Slackline L1 (BAL03) & 140 & 140 & 59/81 \\
Slackline L6 (BAL04) & 139 & 139 & 57/82 \\
Slackline L11 (BAL05) & 138 & 138 & 58/80 \\
Slackline fusion & Not applicable & 132 & 57/75 \\
\bottomrule
\end{tabularx}
\end{table}

Inherited preprocessing independently filtered continuous blocks at 0.5-40 Hz, harmonised sampling to 128 Hz, interpolated approved channels and applied average referencing. Gap-aware epochs overlapped by 50\% within blocks, avoiding exclusion annotations. Features from each epoch remained linked to the corresponding participant, recording and task. Inputs were reused unchanged, with learned preparation and selection restricted to training participants.

\subsection{Feature views, durations and training-fold selection}
The 115 feature definitions in Table~\ref{tab:families} produce individual measurements across channels, channel pairs and other indices; they do not represent the final number of model inputs. Each resulting measurement is a scientific input, distinct from a technical missingness indicator. Family-level duration support differs, and some definitions have additional short-epoch restrictions.

\begin{table}[!htbp]
\centering
\small
\caption{Feature definitions and supported epoch durations.}
\label{tab:families}
\begin{tabularx}{\linewidth}{@{}>{\raggedright\arraybackslash}X lrl@{}}
\toprule
Family & Abbreviation & \shortstack{Feature\\definitions} & Durations \\
\midrule
Temporal/statistical & TEMP & 17 & 2s, 4s, 8s, 16s \\
Conventional spectral & SPEC & 18 & 2s, 4s, 8s, 16s \\
Aperiodic spectral & APER & 18 & 8s, 16s \\
Complexity/fractal & CPLX & 11 & 2s, 4s, 8s, 16s \\
Functional connectivity & CONN & 26 & 2s, 4s, 8s, 16s \\
Graph/network & GRAPH & 15 & 2s, 4s, 8s, 16s \\
Time-frequency decomposition & TFREQ & 10 & 2s, 4s, 8s, 16s \\
\midrule
Total & -- & 115 & Family-dependent \\
\bottomrule
\end{tabularx}
\end{table}
\FloatBarrier

A view comprised one family or a specified combination, using Table~\ref{tab:families}'s abbreviations. Spectral profile (SPEC-P) combined SPEC and APER. Scalar (SCALAR) combined TEMP, SPEC, APER and CPLX. Network (NET) combined CONN and GRAPH, and all-handcrafted (ALL) included all seven families. Combinations required durations supported by every included family, restricting aperiodic-containing views to 8s and 16s.

Descriptive reporting used 8s, supported by all 11 views, while the benchmark retained every supported duration.

Preparation addressed availability and redundancy before selection, using training-participant epoch medians to limit recording-length effects. Summaries required at least ten valid epochs and 50\% valid coverage, with 80\% finite training-participant coverage per input. Constant and duplicate inputs were removed. Training medians supplied imputation values and scaling centres. Scaling used the interquartile range divided by 1.349, falling back to 1.4826 times the median absolute deviation, then sample standard deviation.

Scientific inputs with absolute Spearman correlation at least 0.95 were reduced to deterministic within-block representatives. Family-balanced F ranking allocated two eligible inputs per family before balanced filling, retaining up to 64 scientific inputs in total per fitted view. Eligible missingness indicators with training prevalence between 1\% and 95\% were additional. All operations used training folds, excluding validation/test influence. The learned preparation rules and selected inputs were then applied to individual epochs for classifier training. Separate feature-structure analyses and training-fold selection frequencies characterised inputs independently of held-out attribution.

\subsection{Classifiers and nested participant-held-out evaluation}
The six classifiers cover different modelling assumptions, including two Support Vector Machine (SVM) variants (Table~\ref{tab:classifiers}).

\begin{table}[htbp]
\centering
\small
\caption{Evaluated classifiers and abbreviations.}
\label{tab:classifiers}
\begin{tabularx}{\linewidth}{@{}>{\raggedright\arraybackslash}p{31mm} >{\raggedright\arraybackslash}X l@{}}
\toprule
Modelling group & Classifier & Abbreviation \\
\midrule
Support vector machines & Linear support vector machine & LSVM \\
 & Radial-basis-function support vector machine & RBF-SVM \\
\midrule
Tree ensembles & Random forest & RF \\
 & Histogram gradient boosting & HGB \\
 & XGBoost \cite{chen2016xgboost} & XGB \\
\midrule
Neural network & Compact multilayer perceptron & MLP \\
\bottomrule
\end{tabularx}
\end{table}

Configurations combined cohort, view, duration and classifier: 36 supported view-duration combinations yielded 216 configurations per cohort and 1,080 overall. Outcome-stratified participant partitions comprised three repetitions of five outer folds, each with four inner folds for candidate selection. The 16,200 outer evaluations were not independent participant samples. Participant epochs stayed together, with shared within-cohort partitions and separate participant partitions for fusion. Nesting separated selection from evaluation \cite{varma}.

Within inner-training partitions, preparation, selection and transformation preceded classifier fitting. Predefined classifier settings were evaluated with either no transformation or a Yeo-Johnson transformation. Epoch-level classifier training used inverse participant epoch-count weights before class weighting. Candidates required finite predictions and no convergence warnings in all four inner folds. Mean inner-fold participant AUROC, based on participant-mean raw scores, determined ranking, with ties resolved by average precision, model complexity and deterministic ordering. LSVM and RBF-SVM supplied decision scores, other models native positive-class probabilities.

For each task, an unweighted logistic sigmoid with an intercept and inverse regularisation strength $C=10^6$ was fitted to the selected candidate's pooled inner-validation participant-mean scores. Preparation, transformation and classifier were refitted on outer-training data. The fitted calibration function transformed held-out epoch scores before participant averaging. Calibration thus fitted participant-mean scores but was applied to epoch scores.

Fusion used separate task models sharing a candidate specification. Tuning pooled participant-task scores within inner folds before calibration and fusion. Late fusion equally averaged the three participant-task probabilities. Training-prevalence and stratified-dummy controls remained separate from the evaluated feature-model configurations.

\subsection{Metrics and exploratory statistical comparisons}
Participant-level AUROC was primary. Metrics were recalculated after pooling the five disjoint outer test folds within each repetition, and configuration summaries averaged three repetition metrics. Descriptive medians summarised configuration means across specified sets of evaluated configurations. Accuracy, balanced accuracy, precision, recall, specificity and $F_1$ used a 0.5 probability threshold, with ADHD recruitment positive. $F_1$ was the harmonic mean of participant-level precision and recall within repetitions, then averaged. With no positive predictions, precision and $F_1$ were stored as zero and precision flagged undefined. Brier score measured unthresholded probability error, not complete calibration.

For each task/classifier, the main-text tables selected the highest unrounded mean AUROC across 11 views at 8s after inspection. All metrics came from that row. Different views preclude a controlled classifier-only comparison. Appendix matrices retain every individual-task configuration at 8s.

Per-configuration 95\% intervals used 1,000 outcome-stratified participant-cluster bootstrap draws, retaining all three predictions per sampled participant. Paired comparisons used 20,000 such draws with shared participant multiplicities across configurations. Factor-level means equally averaged configuration AUROCs by view, duration or classifier, distinct from descriptive medians. View/classifier contrasts used common 8s/16s support and duration contrasts seven views available at all four durations. References were SPEC, LSVM and 8s, respectively.

Approximate two-sided null-centred paired bootstrap tests assessed zero factor-level AUROC differences. Holm adjustment covered 72 individual-task contrasts jointly, 18 per task, separately from 18 within-fusion contrasts. Adjusted values, $p_{\mathrm{adj}}$, used a 0.05 threshold. The 95\% intervals were pointwise, not multiplicity-adjusted. Epochs, folds, repetitions and configurations were not independent participants. References and testing families were selected after reviewing results. Fixed-prediction inference excluded uncertainty from refitting, selection and external-population generalisation.

\subsection{Held-out attribution and selection stability}
Feature-view performance does not identify individual contributors. We reconstructed eight 8s configurations selected after predictive inspection, covering 120 outer evaluations, without a new search, and verified that held-out predictions matched the original benchmark predictions. ALL/XGB supplied a common model, view and duration across four tasks without controlling cohort, device or partition differences. In Robots, CONN/XGB examined the observed maximum and NET/XGB a related representation. CONN/RF and CONN/HGB examined recurrence across procedures, not necessarily the next-best classifiers. All explained fits retained 64 scientific inputs.

Tree ensembles permitted a common interventional TreeSHAP (Shapley Additive Explanations for trees) approach, without implying that omitted classes were inferior. TreeSHAP attributes fitted predictions to inputs relative to a background \cite{trees}, here on the model's positive-class probability before external sigmoid calibration. The background sampled one observed epoch per outer-training participant. Every held-out epoch was explained. Signed contributions were averaged within participant before conversion to magnitudes, weighting participants equally within repetition and repetitions equally.

Leading scientific candidates used unconditional mean absolute participant contributions within each procedure, including zeros for available-unselected inputs and excluding unavailable inputs. A scientific input was considered available when present in the original feature view, regardless of whether it survived subsequent training-fold filtering. Conditional-selected means used participants from fits retaining the input and equally weighted available repetitions. Unlike attribution to fitted predictions, selection frequency records retention across fits. Frequency and repetition-level attribution ranks described stability, not additional predictive tests. Technical indicators were distinct from physiological measurements and available only where retained, but included in attribution ranks and top-ten counts.

\section{Results}
We first report predictive performance for selected 8s configurations, then compare feature views, durations and classifiers within tasks and examine which inputs recurred in the held-out explanations. Complete individual-task 8s matrices appear in Tables~\ref{tab:appendix8s-bal02}-\ref{tab:appendix8s-bal05}.

\subsection{Predictive performance across task contexts}
Discrimination was stronger in Robots and weaker across Slackline (Tables~\ref{tab:best8s-bal02}-\ref{tab:best8s-bal05}). These selected 8s configurations are descriptive, not validated optima across durations or prospective model-selection performance. Fixed-prediction AUROC intervals do not correct maximum selection. Bold marks the highest unrounded AUROC, not significance. Matched paired comparisons follow in Sections~\ref{sec:feature-view-findings} and~\ref{sec:duration-classifier-comparisons}. Labels follow Tables~\ref{tab:families}-\ref{tab:classifiers} and the combined-view definitions. CI denotes confidence interval, and Acc., Prec. and Spec. denote accuracy, precision and specificity.

\begin{table}[!htbp]
\centering
\small
\setlength{\tabcolsep}{3pt}
\caption{Predictive performance in Robots using selected feature views at 8s.}
\label{tab:best8s-bal02}
\begin{tabular*}{\linewidth}{@{\extracolsep{\fill}}llrrrrrr@{}}
\toprule
Classifier & Feature view & AUROC [95\% CI] & Acc. & Prec. & Recall & Spec. & $F_1$ \\
\midrule
LSVM & CONN & 0.746 [0.660-0.832] & 0.710 & 0.689 & 0.549 & 0.824 & 0.611 \\
RBF-SVM & NET & 0.728 [0.643-0.816] & 0.683 & 0.649 & 0.510 & 0.806 & 0.570 \\
RF & CONN & 0.772 [0.689-0.855] & 0.734 & 0.735 & 0.562 & 0.856 & 0.637 \\
HGB & CONN & 0.767 [0.684-0.849] & 0.734 & 0.740 & 0.556 & 0.861 & 0.634 \\
MLP & CONN & 0.742 [0.658-0.832] & 0.710 & 0.684 & 0.562 & 0.815 & 0.617 \\
XGB & CONN & \textbf{0.773} [0.692-0.854] & 0.734 & 0.740 & 0.556 & 0.861 & 0.634 \\
\bottomrule
\end{tabular*}
\end{table}

\begin{table}[!htbp]
\centering
\small
\setlength{\tabcolsep}{3pt}
\caption{Predictive performance in Slackline L1 using selected feature views at 8s.}
\label{tab:best8s-bal03}
\begin{tabular*}{\linewidth}{@{\extracolsep{\fill}}llrrrrrr@{}}
\toprule
Classifier & Feature view & AUROC [95\% CI] & Acc. & Prec. & Recall & Spec. & $F_1$ \\
\midrule
LSVM & CONN & 0.528 [0.474-0.582] & 0.574 & 0.306 & 0.034 & 0.967 & 0.058 \\
RBF-SVM & SPEC-P & 0.545 [0.486-0.605] & 0.600 & 0.889 & 0.062 & 0.992 & 0.115 \\
RF & SCALAR & 0.590 [0.502-0.668] & 0.579 & 0.509 & 0.130 & 0.905 & 0.207 \\
HGB & SCALAR & \textbf{0.617} [0.537-0.691] & 0.579 & 0.500 & 0.169 & 0.877 & 0.253 \\
MLP & TFREQ & 0.567 [0.493-0.642] & 0.590 & 0.544 & 0.186 & 0.885 & 0.275 \\
XGB & SCALAR & 0.617 [0.535-0.694] & 0.593 & 0.543 & 0.186 & 0.889 & 0.277 \\
\bottomrule
\end{tabular*}
\end{table}

\begin{table}[!htbp]
\centering
\small
\setlength{\tabcolsep}{3pt}
\caption{Predictive performance in Slackline L6 using selected feature views at 8s.}
\label{tab:best8s-bal04}
\begin{tabular*}{\linewidth}{@{\extracolsep{\fill}}llrrrrrr@{}}
\toprule
Classifier & Feature view & AUROC [95\% CI] & Acc. & Prec. & Recall & Spec. & $F_1$ \\
\midrule
LSVM & SPEC-P & 0.514 [0.456-0.574] & 0.578 & 0.150 & 0.012 & 0.972 & 0.022 \\
RBF-SVM & SPEC & 0.546 [0.480-0.608] & 0.583 & 0.463 & 0.088 & 0.927 & 0.142 \\
RF & SPEC & 0.542 [0.466-0.614] & 0.585 & 0.444 & 0.035 & 0.967 & 0.064 \\
HGB & SPEC & 0.581 [0.504-0.656] & 0.597 & 0.679 & 0.082 & 0.955 & 0.140 \\
MLP & SPEC & \textbf{0.598} [0.520-0.673] & 0.578 & 0.479 & 0.105 & 0.907 & 0.161 \\
XGB & SPEC & 0.574 [0.497-0.655] & 0.590 & 0.518 & 0.082 & 0.943 & 0.139 \\
\bottomrule
\end{tabular*}
\end{table}

\begin{table}[!htbp]
\centering
\small
\setlength{\tabcolsep}{3pt}
\caption{Predictive performance in Slackline L11 using selected feature views at 8s.}
\label{tab:best8s-bal05}
\begin{tabular*}{\linewidth}{@{\extracolsep{\fill}}llrrrrrr@{}}
\toprule
Classifier & Feature view & AUROC [95\% CI] & Acc. & Prec. & Recall & Spec. & $F_1$ \\
\midrule
LSVM & CPLX & \textbf{0.614} [0.527-0.696] & 0.582 & 0.513 & 0.213 & 0.850 & 0.298 \\
RBF-SVM & CPLX & 0.555 [0.488-0.627] & 0.560 & 0.391 & 0.098 & 0.896 & 0.155 \\
RF & CPLX & 0.557 [0.476-0.641] & 0.556 & 0.425 & 0.138 & 0.858 & 0.207 \\
HGB & CPLX & 0.565 [0.489-0.641] & 0.548 & 0.391 & 0.121 & 0.858 & 0.182 \\
MLP & SCALAR & 0.546 [0.476-0.616] & 0.577 & 0.495 & 0.184 & 0.862 & 0.258 \\
XGB & CPLX & 0.594 [0.512-0.675] & 0.556 & 0.434 & 0.155 & 0.846 & 0.227 \\
\bottomrule
\end{tabular*}
\end{table}

\FloatBarrier

CONN was selected for five Robots classifiers and NET for RBF-SVM, whereas selected views differed across Slackline levels. Slackline recall and $F_1$ remained low despite often high specificity. The Robots CONN configuration with XGB also had balanced accuracy 0.708 (95\% CI 0.639-0.785) and Brier score 0.187 (95\% CI 0.157-0.214). Differences between tasks remain descriptive because cohorts, devices and partitions differ.

In Slackline L1, HGB and XGB both round to AUROC 0.617, but the former has the higher unrounded mean. The Slackline L1 and L6 LSVM rows each include one of three repetitions with no positive predictions. Their precision and $F_1$ means include the zeros assigned when no participant was predicted positive, with precision flagged undefined in those repetitions.

Across all 216 configuration means per cohort, median AUROC (IQR, 25th-75th percentiles) was 0.601 (0.556-0.666) for Robots, 0.518 (0.492-0.551) for Slackline L1, 0.487 (0.473-0.512) for Slackline L6, 0.488 (0.466-0.511) for Slackline L11 and 0.477 (0.451-0.507) for fusion. These ranges describe variation across configuration means, not participant-level uncertainty. The summaries cover configurations at all supported durations, rather than the selected 8s subset.

\subsection{Feature-view findings}
\label{sec:feature-view-findings}
\begin{figure}[H]
\centering
\includegraphics[width=.90\linewidth]{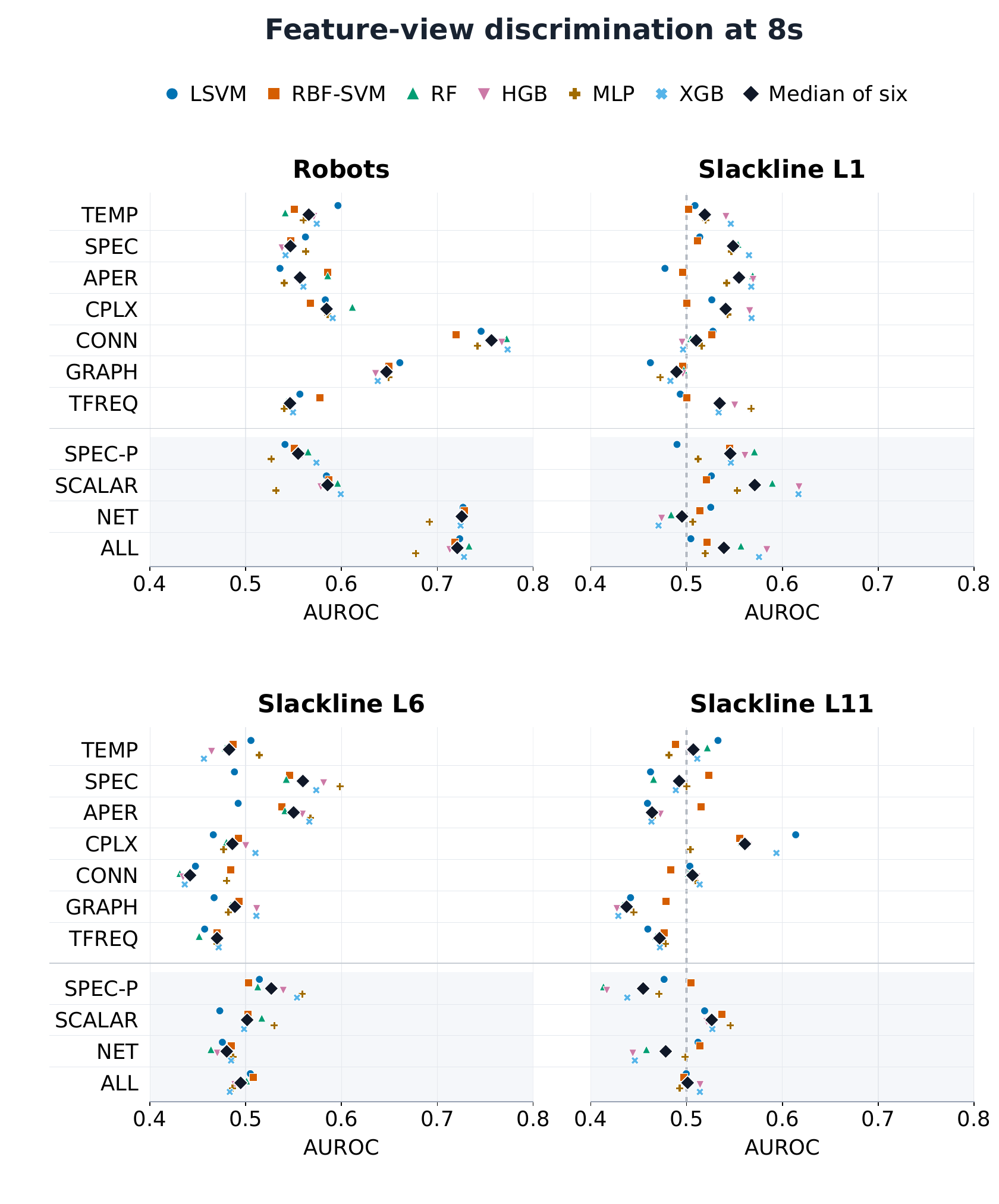}
\caption{Classifier-specific AUROC and feature-view medians across four tasks at 8s.}
\label{fig:views8s}
\end{figure}

At 8s, connectivity-containing views were favourable in Robots, while Slackline's leading views differed (Figure~\ref{fig:views8s}, Table~\ref{tab:views8s}). Coloured symbols show classifier configuration means across three repetitions, with black diamonds marking medians across six classifiers. The dashed line marks AUROC 0.5 and shading combined views. Table cells give these medians plus the separate fusion cohort. These differ from summaries across all supported durations.

\begin{table}[htbp]
\centering
\small
\caption{Median AUROC across classifiers for each feature view and task context at 8s.}
\label{tab:views8s}
\begin{tabularx}{\linewidth}{@{}>{\raggedright\arraybackslash}X rrrrr@{}}
\toprule
Feature view & Robots & \multicolumn{4}{c}{Slackline} \\
\cmidrule(lr){3-6}
 & & L1 & L6 & L11 & Fusion \\
\midrule
\multicolumn{6}{@{}l}{\textit{Individual families}} \\
Temporal/statistical (TEMP) & 0.566 & 0.519 & 0.483 & 0.507 & 0.457 \\
Conventional spectral (SPEC) & 0.547 & 0.549 & 0.560 & 0.492 & 0.482 \\
Aperiodic spectral (APER) & 0.557 & 0.555 & 0.550 & 0.464 & 0.501 \\
Complexity/fractal (CPLX) & 0.584 & 0.541 & 0.486 & 0.561 & 0.545 \\
Functional connectivity (CONN) & 0.756 & 0.510 & 0.442 & 0.506 & 0.480 \\
Graph/network (GRAPH) & 0.647 & 0.489 & 0.489 & 0.437 & 0.428 \\
Time-frequency (TFREQ) & 0.546 & 0.535 & 0.470 & 0.472 & 0.452 \\
\midrule
\multicolumn{6}{@{}l}{\textit{Combined views}} \\
Spectral profile (SPEC-P) & 0.555 & 0.546 & 0.527 & 0.455 & 0.481 \\
Scalar (SCALAR) & 0.585 & 0.571 & 0.502 & 0.526 & 0.561 \\
Network (NET) & 0.726 & 0.495 & 0.480 & 0.478 & 0.464 \\
All handcrafted (ALL) & 0.721 & 0.539 & 0.495 & 0.501 & 0.510 \\
\bottomrule
\end{tabularx}

\end{table}

In the exploratory paired analysis, Robots CONN exceeded the SPEC reference by +0.2083 mean AUROC (95\% CI 0.1076-0.3072, Holm-adjusted $p_{\mathrm{adj}} = 0.0036$). The NET contrast was +0.1828 (95\% CI 0.0766-0.2874, $p_{\mathrm{adj}} = 0.0345$), and the ALL contrast was +0.1692 (95\% CI 0.0886-0.2496, $p_{\mathrm{adj}} = 0.0106$). All intervals are pointwise and $p_{\mathrm{adj}}$ values are Holm-adjusted. These contrasts averaged six classifiers and both 8s and 16s epochs, unlike the table's 8s medians.

\FloatBarrier

\subsection{Duration and classifier comparisons}
\label{sec:duration-classifier-comparisons}
Within-task paired factor-level AUROC differences averaged seven views and six classifiers for duration contrasts, or 11 views at 8s/16s for classifier contrasts. Table~\ref{tab:paired-duration-classifier} reports differences from 8s or LSVM, followed by $p_{\mathrm{adj}}$ in parentheses. Positive differences favour the listed level. Full pointwise 95\% CIs and separate fusion results appear in \ref{app:paired-contrasts}.

On the common duration support, Robots' 16s-minus-8s contrast was $-0.0236$ (95\% CI $-0.0351$ to $-0.0121$, Holm-adjusted $p_{\mathrm{adj}} = 0.0106$). Only this duration contrast and the three Robots view contrasts crossed the corrected threshold. All remaining reference contrasts, including those for Slackline, classifiers against LSVM and within fusion, did not. Non-significance does not establish equivalence.

\begin{table}[!htbp]
\centering
\small
\setlength{\tabcolsep}{3pt}
\caption{Paired duration and classifier contrasts across task contexts.}
\label{tab:paired-duration-classifier}
\begin{tabular*}{\linewidth}{@{\extracolsep{\fill}}lrrrr@{}}
\toprule
Comparison & Robots & Slackline L1 & Slackline L6 & Slackline L11 \\
\midrule
\multicolumn{5}{@{}l}{\textit{Duration versus 8s}} \\
2s & $+0.0082$ ($1$) & $+0.0014$ ($1$) & $+0.0079$ ($1$) & $-0.0036$ ($1$) \\
4s & $-0.0064$ ($1$) & $+0.0024$ ($1$) & $+0.0014$ ($1$) & $0.0000$ ($1$) \\
16s & $-0.0236$ ($0.0106$) & $-0.0195$ ($0.4785$) & $-0.0057$ ($1$) & $+0.0055$ ($1$) \\
\midrule
\multicolumn{5}{@{}l}{\textit{Classifier versus LSVM}} \\
RBF-SVM & $+0.0009$ ($1$) & $+0.0140$ ($1$) & $+0.0073$ ($1$) & $+0.0062$ ($1$) \\
RF & $+0.0053$ ($1$) & $+0.0319$ ($1$) & $+0.0022$ ($1$) & $-0.0131$ ($1$) \\
HGB & $-0.0139$ ($1$) & $+0.0416$ ($0.0612$) & $+0.0092$ ($1$) & $-0.0163$ ($1$) \\
MLP & $-0.0232$ ($1$) & $+0.0242$ ($1$) & $+0.0225$ ($1$) & $-0.0040$ ($1$) \\
XGB & $-0.0030$ ($1$) & $+0.0361$ ($0.3618$) & $+0.0089$ ($1$) & $-0.0065$ ($1$) \\
\bottomrule
\end{tabular*}
\end{table}

\FloatBarrier

\subsection{Held-out attribution and selection stability}
\begin{figure}[H]
\centering
\includegraphics[width=.95\linewidth]{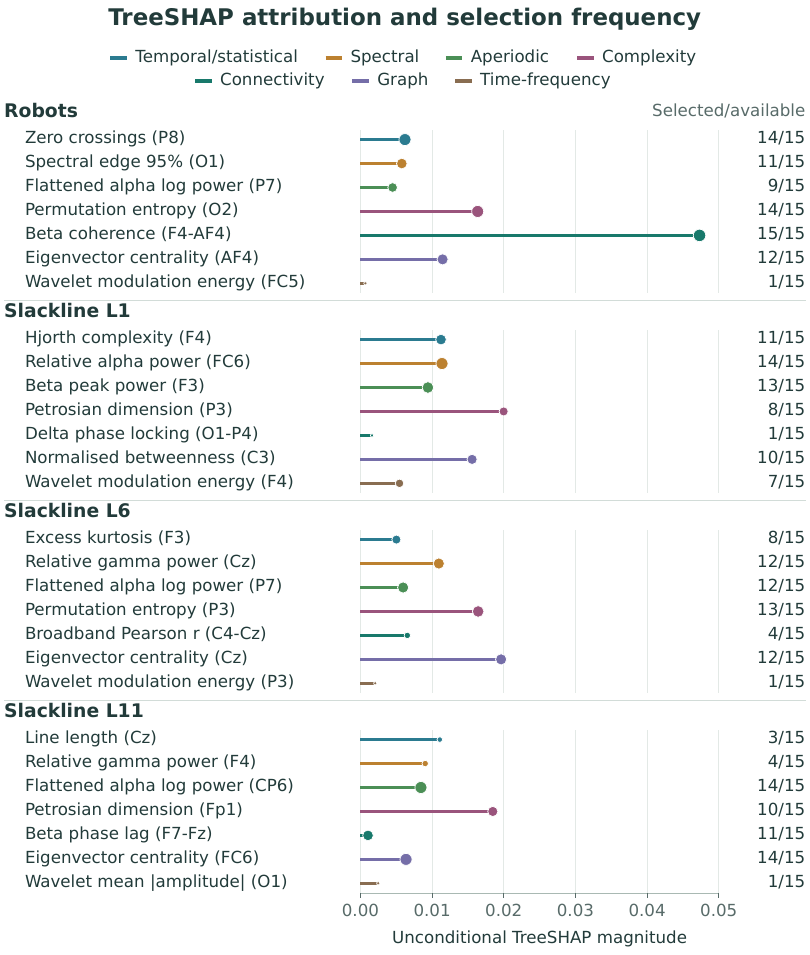}
\caption{Held-out TreeSHAP attribution and selection frequency for the four common ALL/XGB/8s configurations.}
\label{fig:treeshap-selection}
\end{figure}

Attribution and selection frequency describe complementary aspects of fitted-model use. Figure~\ref{fig:treeshap-selection} shows each family's leading scientific input by unconditional attribution in the four common ALL/XGB/8s configurations, an exploratory subset of the eight explained configurations. It neither ranks total family importance nor establishes cross-task superiority. Horizontal position shows unconditional mean absolute participant attribution on the model's positive-class probability before external calibration. Colour denotes family, dot area selection frequency, and counts selected/available fits. Fifteen fits are three repetitions of five folds, not independent participants. Aggregation, weighting and availability follow Section~3.5.

\pagebreak

Beta-band magnitude-squared coherence at F4-AF4 led the repetition-level attribution summaries in all five selected Robots procedures. It was selected and in the top ten in all 15 fits per procedure, although individual-fit ranks varied. This recurrence covered related connectivity-containing views and three tree-model families.

The leading scientific Slackline inputs were less consistently retained and had variable repetition rankings. Petrosian fractal dimension at P3 was selected in 8/15 available fits for L1, and eigenvector centrality at Cz in the positive broadband Pearson-correlation graph in 12/15 for L6. Petrosian fractal dimension at Fp1 was selected in 10/15 available fits for L11. All 120 evaluations passed participant-prediction reproduction and native-probability additivity checks, establishing computational consistency.

\section{Discussion}
The within-task comparisons identify choices worth replicating. The weaker Slackline results limit the case for a single recommendation across task contexts.

\subsection{Representation choices across task contexts}
For RQ1, the clearest within-task differences were in Robots, where connectivity-containing views showed stronger discrimination than the conventional spectral view. Paired contrasts extend the descriptive pattern beyond one favourable configuration, but do not rank CONN, NET and ALL against each other, isolate family effects or establish benefit from adding graph features.

Across task contexts (RQ2), discrimination was strongest in Robots. The leading view varied across the three Slackline levels, and the median AUROC across configurations was near chance at each level. Cohorts, hardware and partitions differed across tasks, so these descriptive comparisons cannot isolate an effect of task design.

\subsection{Epoch duration and classifier choices}
For RQ3, the matched-view comparison in Robots favoured 8s over 16s epochs on average, without establishing a universally optimal duration. The 8s Robots connectivity configuration with XGB had the highest observed mean AUROC, but the paired comparisons against LSVM did not establish a general classifier winner. Replication should therefore vary feature view and epoch duration explicitly to see whether the Robots pattern recurs.

\subsection{Attribution and selection stability}
In the exploratory held-out explanations (RQ4), F4-AF4 beta-band coherence recurred across the five selected Robots procedures. This narrows the replication priority to an individual measurement. Repeated selection and high attribution rank provide complementary information: an input can have a large attribution when selected but appear in few fits. Reporting availability, unconditional importance and selection frequency makes this distinction visible, particularly for the less consistent Slackline inputs.

Attribution describes fitted-model use, not the direction of an ADHD-comparison physiological difference or an independent predictive effect. Correlated inputs can exchange attribution, and the background defines the comparison. Reproduction and additivity checks establish computational consistency, not biological validity.

Other interpretation studies address different questions. Concept sensitivity in EEG abnormality detection \cite{brenner} differs from contributions of engineered inputs to recruitment-group predictions. Shapley-guided reduced-electrode motor-imagery evaluation \cite{perez} tests a downstream benefit not examined here. Our summaries identify measurements for replication, without demonstrating an electrode-reduction benefit.

This addresses the feature-comparison gap \cite{features}, but remains short of biomarker validation \cite{biomarkers}. Independent studies should evaluate connectivity-containing Robots views and recurring coherence while checking whether weaker Slackline patterns persist. Predictive performance and attribution stability require separate assessment because agreement among selected-model explanations is not independent predictive confirmation.

\subsection{Limitations}
One source population supplied overlapping task cohorts. Recruitment group differs from assessment-derived status, precluding diagnostic-validity claims. Devices and separate fusion partitions prevent causal task comparisons and controlled fusion-benefit claims. Estimates are internally held out, without tested cross-task transfer or external validation.

The maximum was selected from a large grid. Exploratory paired intervals condition on fixed predictions, excluding full refitting variability. Explanation configurations were selected after reviewing results. Native-probability attribution does not explain the externally calibrated score, and technical indicators may encode availability rather than physiology.

Background sensitivity covered five Robots pilots, not every explained fit or Slackline. Their agreement cannot establish stability across all backgrounds or task contexts.

\section{Conclusions}
This benchmark provides comparative evidence for choosing EEG representations in paediatric ADHD research. Connectivity-containing Robots views merit independent replication relative to the spectral reference. Their recurring F4-AF4 beta-band coherence input supplies a more specific candidate measurement. Weaker Slackline discrimination and less consistent selection limit generalisation across task contexts. The duration contrast supports evaluating epoch length explicitly, while the classifier comparisons establish no general winner. Future work should test whether these predictive and explanatory patterns recur in independent cohorts under controlled acquisition and evaluation. The present findings guide research priorities without establishing disease mechanisms or clinical diagnostic validity.

\section*{Ethics statement}
This study analysed existing public, anonymised EEG data. It involved no new participant recruitment, intervention or data collection.

\section*{Data and code availability}
The original BALLADEER recordings are available from Figshare \cite{source}. The curated EEG and feature dataset is available from Zenodo under CC BY 4.0 \cite{sanchis2026taskeegdata}. Its preparation software covers EEG quality control, preprocessing and feature extraction and is available under BSD 3-Clause at \url{https://doi.org/10.5281/zenodo.21982948}. The portable research implementation, Docker interface, and aggregate benchmark, statistical and explainability results are available from Zenodo, version 1.0.0 \cite{sanchis2026eeganalysis}. Software is licensed under BSD-3-Clause, while aggregate results and release documentation are licensed under CC BY 4.0.

\section*{Acknowledgements}
This work was supported by Subvenciones para estancias de personal investigador doctor en centros de investigaci\'on radicados fuera de la Comunitat Valenciana para el ejercicio 2026 (CIBEST-2026), grant CIBEST/2026/181, Conselleria d'Educaci\'o, Cultura i Universitats. It was also supported by the BALIDA-AA project (CIPROM/2024/13), funded by Conselleria de Educaci\'on, Cultura, Universidades y Empleo (Generalitat Valenciana). Additional support was provided by the KOSMOS-UA project (PID2024-155363OB-C43), funded by the Spanish Ministry of Science and Innovation. The funders had no role in study design, analysis, writing or the decision to submit.

\section*{Declaration of competing interests}
The authors declare no competing interests.

\section*{CRediT authorship contribution statement}
\textbf{Javier Sanchis}: Conceptualization, Data curation, Formal analysis, Investigation, Methodology, Resources, Software, Supervision, Validation, Visualization, Writing - original draft, Writing - review \& editing.

\textbf{Jorge Garc\'ia-Carrasco} and \textbf{\'Alvaro Navarro}: Conceptualization, Data curation, Formal analysis, Investigation, Methodology, Resources, Validation, Visualization, Writing - original draft, Writing - review \& editing.

\textbf{Tahar Kechadi} and \textbf{Juan Trujillo}: Conceptualization, Formal analysis, Funding acquisition, Project administration, Supervision, Writing - review \& editing.

\section*{Declaration of generative AI and AI-assisted technologies in the manuscript preparation process}
During the preparation of this work the authors used ChatGPT to polish and rephrase sentences. After using this tool/service, the authors reviewed and edited the content as needed and take full responsibility for the published article.

\PreprintBibliography{references}
\clearpage

\appendix
\renewcommand{\thesection}{Appendix~\Alph{section}}
\renewcommand{\thetable}{\Alph{section}.\arabic{table}}
\section{Complete 8s feature-view comparisons}
\label{app:8s-matrices}
\setcounter{table}{0}
\renewcommand{\theHtable}{appendix.\arabic{table}}

The complete matrices show the seven families and four combined views for each classifier within each individual task (Tables~\ref{tab:appendix8s-bal02}-\ref{tab:appendix8s-bal05}). Each cell is the participant-level mean AUROC across three repetitions for one fixed 8s configuration, rounded to three decimal places. All 11 views are retained rather than selecting a maximum. These descriptive values are distinct from the medians across classifiers in Table~\ref{tab:views8s} and do not establish significant differences or causal task effects. Feature-view and classifier abbreviations follow Tables~\ref{tab:families} and~\ref{tab:classifiers} and the combined-view definitions in Methods.

\begin{table}[!htbp]
\centering
\small
\setlength{\tabcolsep}{3pt}
\caption{Mean AUROC at 8s for Robots.}
\label{tab:appendix8s-bal02}
\begin{tabularx}{\linewidth}{@{}>{\raggedright\arraybackslash}X rrrrrr@{}}
\toprule
Feature view & LSVM & RBF-SVM & RF & HGB & MLP & XGB \\
\midrule
\multicolumn{7}{@{}l}{\textit{Individual families}} \\
Temporal/statistical (TEMP) & 0.596 & 0.551 & 0.541 & 0.571 & 0.560 & 0.574 \\
Conventional spectral (SPEC) & 0.562 & 0.547 & 0.546 & 0.538 & 0.563 & 0.542 \\
Aperiodic spectral (APER) & 0.536 & 0.586 & 0.586 & 0.553 & 0.540 & 0.560 \\
Complexity/fractal (CPLX) & 0.583 & 0.567 & 0.611 & 0.583 & 0.586 & 0.591 \\
Functional connectivity (CONN) & 0.746 & 0.720 & 0.772 & 0.767 & 0.742 & 0.773 \\
Graph/network (GRAPH) & 0.661 & 0.649 & 0.645 & 0.636 & 0.649 & 0.638 \\
Time-frequency (TFREQ) & 0.557 & 0.577 & 0.543 & 0.543 & 0.540 & 0.549 \\
\midrule
\multicolumn{7}{@{}l}{\textit{Combined views}} \\
Spectral profile (SPEC-P) & 0.541 & 0.551 & 0.565 & 0.559 & 0.527 & 0.574 \\
Scalar (SCALAR) & 0.584 & 0.587 & 0.596 & 0.578 & 0.532 & 0.599 \\
Network (NET) & 0.727 & 0.728 & 0.728 & 0.722 & 0.692 & 0.724 \\
All handcrafted (ALL) & 0.723 & 0.718 & 0.733 & 0.713 & 0.677 & 0.728 \\
\bottomrule
\end{tabularx}
\end{table}

\begin{table}[!htbp]
\centering
\small
\setlength{\tabcolsep}{3pt}
\caption{Mean AUROC at 8s for Slackline L1.}
\label{tab:appendix8s-bal03}
\begin{tabularx}{\linewidth}{@{}>{\raggedright\arraybackslash}X rrrrrr@{}}
\toprule
Feature view & LSVM & RBF-SVM & RF & HGB & MLP & XGB \\
\midrule
\multicolumn{7}{@{}l}{\textit{Individual families}} \\
Temporal/statistical (TEMP) & 0.509 & 0.502 & 0.518 & 0.541 & 0.520 & 0.546 \\
Conventional spectral (SPEC) & 0.514 & 0.512 & 0.554 & 0.550 & 0.547 & 0.565 \\
Aperiodic spectral (APER) & 0.478 & 0.496 & 0.569 & 0.569 & 0.542 & 0.568 \\
Complexity/fractal (CPLX) & 0.526 & 0.501 & 0.539 & 0.566 & 0.543 & 0.568 \\
Functional connectivity (CONN) & 0.528 & 0.526 & 0.504 & 0.495 & 0.516 & 0.496 \\
Graph/network (GRAPH) & 0.462 & 0.496 & 0.497 & 0.496 & 0.473 & 0.483 \\
Time-frequency (TFREQ) & 0.493 & 0.500 & 0.536 & 0.550 & 0.567 & 0.534 \\
\midrule
\multicolumn{7}{@{}l}{\textit{Combined views}} \\
Spectral profile (SPEC-P) & 0.490 & 0.545 & 0.571 & 0.561 & 0.512 & 0.546 \\
Scalar (SCALAR) & 0.526 & 0.521 & 0.590 & 0.617 & 0.553 & 0.617 \\
Network (NET) & 0.525 & 0.514 & 0.484 & 0.474 & 0.507 & 0.471 \\
All handcrafted (ALL) & 0.504 & 0.521 & 0.557 & 0.584 & 0.520 & 0.576 \\
\bottomrule
\end{tabularx}
\end{table}

\begin{table}[!htbp]
\centering
\small
\setlength{\tabcolsep}{3pt}
\caption{Mean AUROC at 8s for Slackline L6.}
\label{tab:appendix8s-bal04}
\begin{tabularx}{\linewidth}{@{}>{\raggedright\arraybackslash}X rrrrrr@{}}
\toprule
Feature view & LSVM & RBF-SVM & RF & HGB & MLP & XGB \\
\midrule
\multicolumn{7}{@{}l}{\textit{Individual families}} \\
Temporal/statistical (TEMP) & 0.505 & 0.487 & 0.479 & 0.464 & 0.514 & 0.456 \\
Conventional spectral (SPEC) & 0.488 & 0.546 & 0.542 & 0.581 & 0.598 & 0.574 \\
Aperiodic spectral (APER) & 0.492 & 0.538 & 0.541 & 0.559 & 0.567 & 0.566 \\
Complexity/fractal (CPLX) & 0.466 & 0.492 & 0.480 & 0.500 & 0.477 & 0.510 \\
Functional connectivity (CONN) & 0.448 & 0.485 & 0.431 & 0.435 & 0.480 & 0.436 \\
Graph/network (GRAPH) & 0.467 & 0.493 & 0.484 & 0.511 & 0.482 & 0.511 \\
Time-frequency (TFREQ) & 0.457 & 0.470 & 0.452 & 0.472 & 0.470 & 0.472 \\
\midrule
\multicolumn{7}{@{}l}{\textit{Combined views}} \\
Spectral profile (SPEC-P) & 0.514 & 0.503 & 0.513 & 0.539 & 0.559 & 0.554 \\
Scalar (SCALAR) & 0.473 & 0.502 & 0.517 & 0.501 & 0.530 & 0.498 \\
Network (NET) & 0.476 & 0.485 & 0.464 & 0.470 & 0.487 & 0.485 \\
All handcrafted (ALL) & 0.505 & 0.508 & 0.501 & 0.489 & 0.486 & 0.483 \\
\bottomrule
\end{tabularx}
\end{table}

\begin{table}[!htbp]
\centering
\small
\setlength{\tabcolsep}{3pt}
\caption{Mean AUROC at 8s for Slackline L11.}
\label{tab:appendix8s-bal05}
\begin{tabularx}{\linewidth}{@{}>{\raggedright\arraybackslash}X rrrrrr@{}}
\toprule
Feature view & LSVM & RBF-SVM & RF & HGB & MLP & XGB \\
\midrule
\multicolumn{7}{@{}l}{\textit{Individual families}} \\
Temporal/statistical (TEMP) & 0.533 & 0.488 & 0.522 & 0.503 & 0.482 & 0.511 \\
Conventional spectral (SPEC) & 0.462 & 0.523 & 0.466 & 0.496 & 0.500 & 0.489 \\
Aperiodic spectral (APER) & 0.459 & 0.515 & 0.464 & 0.473 & 0.464 & 0.463 \\
Complexity/fractal (CPLX) & 0.614 & 0.555 & 0.557 & 0.565 & 0.504 & 0.594 \\
Functional connectivity (CONN) & 0.503 & 0.484 & 0.502 & 0.511 & 0.509 & 0.514 \\
Graph/network (GRAPH) & 0.442 & 0.479 & 0.433 & 0.427 & 0.445 & 0.429 \\
Time-frequency (TFREQ) & 0.460 & 0.477 & 0.468 & 0.471 & 0.478 & 0.472 \\
\midrule
\multicolumn{7}{@{}l}{\textit{Combined views}} \\
Spectral profile (SPEC-P) & 0.477 & 0.504 & 0.413 & 0.417 & 0.471 & 0.438 \\
Scalar (SCALAR) & 0.519 & 0.537 & 0.526 & 0.521 & 0.546 & 0.527 \\
Network (NET) & 0.512 & 0.514 & 0.458 & 0.444 & 0.499 & 0.446 \\
All handcrafted (ALL) & 0.500 & 0.497 & 0.503 & 0.514 & 0.493 & 0.514 \\
\bottomrule
\end{tabularx}
\end{table}

\clearpage

\section{Paired duration and classifier comparisons}
\label{app:paired-contrasts}
\setcounter{table}{0}
\renewcommand{\theHtable}{appendixB.\arabic{table}}

The individual-task contrasts summarised in Table~\ref{tab:paired-duration-classifier} are reported with full pointwise 95\% CIs in Tables~\ref{app:paired-bal02}-\ref{app:paired-bal05}. The separate within-fusion comparisons appear in Table~\ref{app:paired-fusion}. Differences subtract the reference (8s or LSVM) from the listed level. Duration contrasts average seven views and six classifiers. Classifier contrasts average 11 views and the common 8s/16s support. All estimates condition on fixed predictions.

Holm adjustment retains the joint family of 72 individual-task contrasts and the separate family of 18 fusion contrasts, including the view contrasts not reproduced here. Pointwise intervals are not adjusted for multiplicity, so an interval excluding zero need not correspond to a significant $p_{\mathrm{adj}}$ value. Adjusted values equal to 1 are reported as 1. Other adjusted values are rounded to four decimal places. The Slackline L11 4s contrast rounds to 0.0000 from a small negative difference, not an exact equality.

\begin{table}[!htbp]
\centering
\small
\setlength{\tabcolsep}{4pt}
\caption{Paired duration and classifier contrasts for Robots.}
\label{app:paired-bal02}
\begin{tabular*}{\linewidth}{@{\extracolsep{\fill}}lrrr@{}}
\toprule
Comparison & $\Delta$AUROC & Pointwise 95\% CI & Holm-adjusted $p$ \\
\midrule
\multicolumn{4}{@{}l}{\textit{Duration versus 8s}} \\
2s & $+0.0082$ & $-0.0212$ to $+0.0375$ & $1$ \\
4s & $-0.0064$ & $-0.0327$ to $+0.0200$ & $1$ \\
16s & $-0.0236$ & $-0.0351$ to $-0.0121$ & $0.0106$ \\
\midrule
\multicolumn{4}{@{}l}{\textit{Classifier versus LSVM}} \\
RBF-SVM & $+0.0009$ & $-0.0173$ to $+0.0195$ & $1$ \\
RF & $+0.0053$ & $-0.0174$ to $+0.0275$ & $1$ \\
HGB & $-0.0139$ & $-0.0387$ to $+0.0102$ & $1$ \\
MLP & $-0.0232$ & $-0.0546$ to $+0.0079$ & $1$ \\
XGB & $-0.0030$ & $-0.0246$ to $+0.0182$ & $1$ \\
\bottomrule
\end{tabular*}
\end{table}

\begin{table}[!htbp]
\centering
\small
\setlength{\tabcolsep}{4pt}
\caption{Paired duration and classifier contrasts for Slackline L1.}
\label{app:paired-bal03}
\begin{tabular*}{\linewidth}{@{\extracolsep{\fill}}lrrr@{}}
\toprule
Comparison & $\Delta$AUROC & Pointwise 95\% CI & Holm-adjusted $p$ \\
\midrule
\multicolumn{4}{@{}l}{\textit{Duration versus 8s}} \\
2s & $+0.0014$ & $-0.0219$ to $+0.0240$ & $1$ \\
4s & $+0.0024$ & $-0.0156$ to $+0.0202$ & $1$ \\
16s & $-0.0195$ & $-0.0339$ to $-0.0052$ & $0.4785$ \\
\midrule
\multicolumn{4}{@{}l}{\textit{Classifier versus LSVM}} \\
RBF-SVM & $+0.0140$ & $-0.0070$ to $+0.0357$ & $1$ \\
RF & $+0.0319$ & $+0.0054$ to $+0.0585$ & $1$ \\
HGB & $+0.0416$ & $+0.0164$ to $+0.0668$ & $0.0612$ \\
MLP & $+0.0242$ & $+0.0032$ to $+0.0452$ & $1$ \\
XGB & $+0.0361$ & $+0.0108$ to $+0.0618$ & $0.3618$ \\
\bottomrule
\end{tabular*}
\end{table}

\begin{table}[!htbp]
\centering
\small
\setlength{\tabcolsep}{4pt}
\caption{Paired duration and classifier contrasts for Slackline L6.}
\label{app:paired-bal04}
\begin{tabular*}{\linewidth}{@{\extracolsep{\fill}}lrrr@{}}
\toprule
Comparison & $\Delta$AUROC & Pointwise 95\% CI & Holm-adjusted $p$ \\
\midrule
\multicolumn{4}{@{}l}{\textit{Duration versus 8s}} \\
2s & $+0.0079$ & $-0.0164$ to $+0.0322$ & $1$ \\
4s & $+0.0014$ & $-0.0190$ to $+0.0215$ & $1$ \\
16s & $-0.0057$ & $-0.0189$ to $+0.0073$ & $1$ \\
\midrule
\multicolumn{4}{@{}l}{\textit{Classifier versus LSVM}} \\
RBF-SVM & $+0.0073$ & $-0.0143$ to $+0.0291$ & $1$ \\
RF & $+0.0022$ & $-0.0203$ to $+0.0245$ & $1$ \\
HGB & $+0.0092$ & $-0.0139$ to $+0.0319$ & $1$ \\
MLP & $+0.0225$ & $-0.0015$ to $+0.0463$ & $1$ \\
XGB & $+0.0089$ & $-0.0144$ to $+0.0324$ & $1$ \\
\bottomrule
\end{tabular*}
\end{table}

\begin{table}[!htbp]
\centering
\small
\setlength{\tabcolsep}{4pt}
\caption{Paired duration and classifier contrasts for Slackline L11.}
\label{app:paired-bal05}
\begin{tabular*}{\linewidth}{@{\extracolsep{\fill}}lrrr@{}}
\toprule
Comparison & $\Delta$AUROC & Pointwise 95\% CI & Holm-adjusted $p$ \\
\midrule
\multicolumn{4}{@{}l}{\textit{Duration versus 8s}} \\
2s & $-0.0036$ & $-0.0236$ to $+0.0161$ & $1$ \\
4s & $0.0000$ & $-0.0205$ to $+0.0208$ & $1$ \\
16s & $+0.0055$ & $-0.0108$ to $+0.0223$ & $1$ \\
\midrule
\multicolumn{4}{@{}l}{\textit{Classifier versus LSVM}} \\
RBF-SVM & $+0.0062$ & $-0.0169$ to $+0.0295$ & $1$ \\
RF & $-0.0131$ & $-0.0330$ to $+0.0067$ & $1$ \\
HGB & $-0.0163$ & $-0.0356$ to $+0.0029$ & $1$ \\
MLP & $-0.0040$ & $-0.0267$ to $+0.0186$ & $1$ \\
XGB & $-0.0065$ & $-0.0247$ to $+0.0116$ & $1$ \\
\bottomrule
\end{tabular*}
\end{table}

\begin{table}[!htbp]
\centering
\small
\setlength{\tabcolsep}{4pt}
\caption{Paired contrasts within the Slackline fusion cohort.}
\label{app:paired-fusion}
\begin{tabular*}{\linewidth}{@{\extracolsep{\fill}}lrrr@{}}
\toprule
Comparison & $\Delta$AUROC & Pointwise 95\% CI & Holm-adjusted $p$ \\
\midrule
\multicolumn{4}{@{}l}{\textit{Duration versus 8s}} \\
2s & $-0.0137$ & $-0.0372$ to $+0.0092$ & $1$ \\
4s & $-0.0079$ & $-0.0293$ to $+0.0133$ & $1$ \\
16s & $+0.0051$ & $-0.0068$ to $+0.0170$ & $1$ \\
\midrule
\multicolumn{4}{@{}l}{\textit{Classifier versus LSVM}} \\
RBF-SVM & $-0.0044$ & $-0.0333$ to $+0.0255$ & $1$ \\
RF & $-0.0025$ & $-0.0241$ to $+0.0198$ & $1$ \\
HGB & $+0.0010$ & $-0.0235$ to $+0.0260$ & $1$ \\
MLP & $+0.0112$ & $-0.0155$ to $+0.0379$ & $1$ \\
XGB & $+0.0023$ & $-0.0236$ to $+0.0286$ & $1$ \\
\bottomrule
\end{tabular*}
\end{table}

\clearpage


@article{features,
  author  = {Arpaia, Pasquale and Covino, Attilio and Cristaldi, Loredana and Frosolone, Mirco and Gargiulo, Ludovica and Mancino, Francesca and Mantile, Federico and Moccaldi, Nicola},
  title   = {A systematic review on feature extraction in electroencephalography-based diagnostics and therapy in attention deficit hyperactivity disorder},
  journal = {Sensors},
  year    = {2022},
  volume  = {22},
  number  = {13},
  pages   = {4934},
  doi     = {10.3390/s22134934}
}

@article{quinterolopez2025review,
  author  = {Quintero L{\'o}pez, Catalina and Gil Vera, V{\'i}ctor D. and Ruiz Quintero, Mar{\'i}a J.},
  title   = {Diagnosis of {ADHD} in Children with {EEG} and Machine Learning: Systematic Review and Meta-analysis},
  journal = {Clinical and Health},
  year    = {2025},
  volume  = {36},
  number  = {2},
  pages   = {109--121},
  doi     = {10.5093/clh2025a16}
}

@article{donoghue2020,
  author  = {Donoghue, Thomas and Haller, Matar and Peterson, Erik J. and Varma, Paroma and Sebastian, Priyadarshini and Gao, Richard and Noto, Torben and Lara, Antonio H. and Wallis, Joni D. and Knight, Robert T. and Shestyuk, Avgusta and Voytek, Bradley},
  title   = {Parameterizing neural power spectra into periodic and aperiodic components},
  journal = {Nature Neuroscience},
  year    = {2020},
  volume  = {23},
  number  = {12},
  pages   = {1655--1665},
  doi     = {10.1038/s41593-020-00744-x}
}

@article{mohammedatheef2026mrsshap,
  author  = {G A, Mohammed Atheef and G, Shyam Pranav and Powar, Omkar S.},
  title   = {{MRS-SHAP}: Multi-objective explainable {EEG} feature selection framework for {ADHD} detection},
  journal = {Discover Applied Sciences},
  year    = {2026},
  volume  = {8},
  pages   = {764},
  doi     = {10.1007/s42452-026-08788-7}
}

@article{biomarkers,
  author  = {Cortese, Samuele and Solmi, Marco and Michelini, Giorgia and Bellato, Alessio and Blanner, Christina and Canozzi, Andrea and Eudave, Luis and Farhat, Luis C. and H{\o}jlund, Mikkel and K{\"o}hler-Forsberg, Ole and Leffa, Douglas Teixeira and Rohde, Christopher and {de Pablo}, Gonzalo Salazar and Vita, Giovanni and Wesselhoeft, Rikke and Martin, Joanna and Baumeister, Sarah and Bozhilova, Natali S. and Carlisi, Christina O. and Leno, Virginia Carter and Floris, Dorothea L. and Holz, Nathalie E. and Kraaijenvanger, Eline J. and Sacu, Seda and Vainieri, Isabella and Ostuzzi, Giovanni and Barbui, Corrado and Correll, Christoph U.},
  title   = {Candidate diagnostic biomarkers for neurodevelopmental disorders in children and adolescents: a systematic review},
  journal = {World Psychiatry},
  year    = {2023},
  volume  = {22},
  number  = {1},
  pages   = {129--149},
  doi     = {10.1002/wps.21037}
}

@article{balladeer,
  author  = {Trujillo, Juan and Ferrer-Cascales, Rosario and Teruel, Miguel A. and Ruiz-Robledillo, Nicol{\'a}s and Sanchis, Javier and Garc{\'i}a-Ponsoda, Sandra and Panagiotidis-Arrizabalaga, Alejandro and Albaladejo-Bl{\'a}zquez, Natalia and Mart{\'i}nez-Nicol{\'a}s, {\'A}ngela and Garc{\'i}a-Carrasco, Jorge and Reina, Alejandro and Lavalle, Ana and Mat{\'e}, Alejandro and Costa-L{\'o}pez, Borja},
  title   = {A multimodal dataset for neurophysiological and {AI} applications},
  journal = {Scientific Data},
  year    = {2026},
  volume  = {13},
  number  = {1},
  pages   = {436},
  doi     = {10.1038/s41597-026-06758-7}
}

@article{varma,
  author  = {Varma, Sudhir and Simon, Richard},
  title   = {Bias in error estimation when using cross-validation for model selection},
  journal = {BMC Bioinformatics},
  year    = {2006},
  volume  = {7},
  number  = {1},
  pages   = {91},
  doi     = {10.1186/1471-2105-7-91}
}

@article{trees,
  author  = {Lundberg, Scott M. and Erion, Gabriel and Chen, Hugh and DeGrave, Alex and Prutkin, Jordan M. and Nair, Bala and Katz, Ronit and Himmelfarb, Jonathan and Bansal, Nisha and Lee, Su-In},
  title   = {From local explanations to global understanding with explainable {AI} for trees},
  journal = {Nature Machine Intelligence},
  year    = {2020},
  volume  = {2},
  pages   = {56--67},
  doi     = {10.1038/s42256-019-0138-9}
}

@misc{source,
  author    = {Trujillo, Juan and Ferrer-Cascales, Rosario and Teruel, Miguel A. and Ruiz-Robledillo, Nicol{\'a}s and Sanchis, Javier and Garc{\'i}a-Ponsoda, Sandra and Panagiotidis-Arrizabalaga, Alejandro and Albaladejo-Bl{\'a}zquez, Natalia and Mart{\'i}nez-Nicol{\'a}s, {\'A}ngela and Garc{\'i}a-Carrasco, Jorge and Reina, Alejandro and Lavalle, Ana and Mat{\'e}, Alejandro and Costa-L{\'o}pez, Borja},
  title     = {A Multimodal Dataset for Neurophysiological and {AI} Applications},
  howpublished = {Figshare},
  year      = {2026},
  note      = {Version 2, CC BY 4.0},
  doi       = {10.6084/m9.figshare.28676042.v2},
  url       = {https://doi.org/10.6084/m9.figshare.28676042.v2}
}

@article{robots,
 author = {Panagiotidis-Arrizabalaga, Alejandro and Lavalle, Ana and Teruel, Miguel A. and Sanchis, Javier and Ruiz-Robledillo, Nicol{\'a}s and Costa-L{\'o}pez, Borja},
 title = {Serious Game-Based Attention Assessment for Attention-Deficit/Hyperactivity Disorder in Children and Adolescents: Preliminary Cross-Sectional Convergent Validation Study},
 journal = {JMIR Serious Games},
 year = {2026}, volume = {14}, pages = {e80441}, doi = {10.2196/80441}
}

@article{slackline,
 author = {Ruiz-Robledillo, Nicol{\'a}s and Lucas, Ignacio and Ferrer-Cascales, Rosario and Albaladejo-Bl{\'a}zquez, Natalia and Sanchis, Javier and Trujillo, Juan},
 title = {Discriminative Power of the Serious Game Attention Slackline in Children and Adolescents With and Without Attention-Deficit/Hyperactivity Disorder: Validation Study},
 journal = {JMIR Serious Games},
 year = {2025}, volume = {13}, pages = {e65170}, doi = {10.2196/65170}
}

@article{brenner,
 author = {Brenner, Alexander and Knispel, Felix and Fischer, Florian P. and Rossmanith, Peter and Weber, Yvonne and Koch, Henner and R{\"o}hrig, Rainer and Varghese, Julian and Kutafina, Ekaterina},
 title = {Concept-based {AI} interpretability in physiological time-series data: Example of abnormality detection in electroencephalography},
 journal = {Computer Methods and Programs in Biomedicine},
 year = {2024}, volume = {257}, pages = {108448}, doi = {10.1016/j.cmpb.2024.108448}
}

@article{perez,
 author = {P{\'e}rez-Velasco, Sergio and Marcos-Mart{\'i}nez, Diego and Santamar{\'i}a-V{\'a}zquez, Eduardo and Mart{\'i}nez-Cagigal, V{\'i}ctor and Moreno-Calder{\'o}n, Selene and Hornero, Roberto},
 title = {Unraveling motor imagery brain patterns using explainable artificial intelligence based on {Shapley} values},
 journal = {Computer Methods and Programs in Biomedicine},
 year = {2024}, volume = {246}, pages = {108048}, doi = {10.1016/j.cmpb.2024.108048}
}

@article{brookshire2024eegleakage,
  author  = {Brookshire, Geoffrey and Kasper, Jake and Blauch, Nicholas M. and Wu, Yunan Charles and Glatt, Ryan and Merrill, David A. and Gerrol, Spencer and Yoder, Keith J. and Quirk, Colin and Lucero, Ch{\'e}},
  title   = {Data leakage in deep learning studies of translational {EEG}},
  journal = {Frontiers in Neuroscience},
  year    = {2024},
  volume  = {18},
  pages   = {1373515},
  doi     = {10.3389/fnins.2024.1373515}
}

@article{delpup2025partitioning,
  author  = {Del Pup, Federico and Zanola, Andrea and Tshimanga, Louis Fabrice and Bertoldo, Alessandra and Finos, Livio and Atzori, Manfredo},
  title   = {The role of data partitioning on the performance of {EEG}-based deep learning models in supervised cross-subject analysis: A preliminary study},
  journal = {Computers in Biology and Medicine},
  year    = {2025},
  volume  = {196},
  pages   = {110608},
  doi     = {10.1016/j.compbiomed.2025.110608}
}

@article{varoquaux2018crossvalidation,
  author  = {Varoquaux, Ga{\"e}l},
  title   = {Cross-validation failure: Small sample sizes lead to large error bars},
  journal = {NeuroImage},
  year    = {2018},
  volume  = {180},
  pages   = {68--77},
  doi     = {10.1016/j.neuroimage.2017.06.061}
}

@article{garciaponsoda2024refining,
  author  = {Garc{\'i}a-Ponsoda, Sandra and Mat{\'e}, Alejandro and Trujillo, Juan},
  title   = {Refining {ADHD} diagnosis with {EEG}: The impact of preprocessing and temporal segmentation on classification accuracy},
  journal = {Computers in Biology and Medicine},
  year    = {2024},
  volume  = {183},
  pages   = {109305},
  doi     = {10.1016/j.compbiomed.2024.109305}
}

@article{lohani2025machine,
  author  = {Lohani, Dhruv Chandra and Chawla, Vaishali and Rana, Bharti},
  title   = {A systematic literature review of machine learning techniques for the detection of attention-deficit/hyperactivity disorder using {MRI} and/or {EEG} data},
  journal = {Neuroscience},
  year    = {2025},
  volume  = {570},
  pages   = {110--131},
  doi     = {10.1016/j.neuroscience.2025.02.019}
}

@article{safi2026amplitude,
  author  = {Safi, Mehrnoosh Sadat and Safi, Seyed Mohammad Mehdi},
  title   = {Diagnosis of {ADHD} in children from {EEG} signals using amplitude modulation features},
  journal = {Computers in Biology and Medicine},
  year    = {2026},
  volume  = {213},
  pages   = {111828},
  doi     = {10.1016/j.compbiomed.2026.111828}
}

@article{ghaderi2025altered,
  author  = {Ghaderi, Amir Hossein and Taghizadeh, Shiva and Nazari, Mohammad Ali},
  title   = {Altered visual network modularity and communication in {ADHD} subtypes: Classification via source-localized {EEG} modules},
  journal = {Biomedical Signal Processing and Control},
  year    = {2025},
  volume  = {106},
  pages   = {107755},
  doi     = {10.1016/j.bspc.2025.107755}
}

@article{lyu2026eeg,
  author  = {Lyu, RuiFang},
  title   = {Deep learning approaches for {EEG}-based healthcare applications: a comprehensive review},
  journal = {Frontiers in Human Neuroscience},
  year    = {2026},
  volume  = {19},
  pages   = {1689073},
  doi     = {10.3389/fnhum.2025.1689073}
}

@misc{nice2018adhd,
  author       = {{National Institute for Health and Care Excellence}},
  title        = {Attention deficit hyperactivity disorder: diagnosis and management},
  howpublished = {NICE guideline {NG87}},
  year         = {2018},
  url          = {https://www.nice.org.uk/guidance/ng87},
  note         = {Accessed 30 October 2026}
}

@inproceedings{chen2016xgboost,
  author    = {Chen, Tianqi and Guestrin, Carlos},
  title     = {{XGBoost}: A Scalable Tree Boosting System},
  booktitle = {Proceedings of the 22nd ACM SIGKDD International Conference on Knowledge Discovery and Data Mining},
  year      = {2016},
  pages     = {785--794},
  publisher = {Association for Computing Machinery},
  doi       = {10.1145/2939672.2939785}
}

@misc{sanchis2026taskeegdata,
  author = {Sanchis, Javier and Navarro, {\'A}lvaro and Garc{\'i}a-Carrasco, Jorge and Kechadi, Mohand Tahar and Trujillo, Juan},
  title = {{EEG Dataset for ADHD Research in Children and Adolescents: Preprocessed Task Recordings and Seven Families of Aligned Features}},
  howpublished = {Zenodo},
  year = {2026},
  note = {Dataset, version 1.0.0},
  doi = {10.5281/zenodo.21982978},
  url = {https://doi.org/10.5281/zenodo.21982978}
}

@misc{sanchis2026eeganalysis,
  author = {Sanchis, Javier and Garc{\'i}a-Carrasco, Jorge and Navarro, {\'A}lvaro and Kechadi, Mohand Tahar and Trujillo, Juan},
  title = {Benchmarking {EEG} feature families for {ADHD} research in children and adolescents: performance and explainability across attention tasks: {Source Code}},
  howpublished = {Zenodo},
  year = {2026},
  note = {Computer software, version 1.0.0},
  doi = {10.5281/zenodo.23884319},
  url = {https://doi.org/10.5281/zenodo.23884319}
}
\end{document}